\documentclass[sigconf]{acmart}
\usepackage{booktabs} 
\usepackage[ruled]{algorithm2e} 
\usepackage{multirow}
\usepackage{subfigure}
\usepackage{float}

\begin{document}
\title{GANE: A Generative Adversarial Network Embedding}
\subtitle{}

\author{
Huiting~Hong,
Xin~Li
and Mingzhong~Wang
}

\begin{abstract}
Network embedding has become a hot research topic recently which can provide low-dimensional feature representations for many machine learning applications. Current work focuses on either (1)
whether the embedding is designed as an unsupervised learning task by explicitly preserving the structural connectivity in the network, or (2) whether the embedding is a by-product during the supervised learning of a specific discriminative task in a deep neural network. In this paper, we focus on bridging the gap of the two lines of the research. We propose to adapt the Generative Adversarial model to perform network embedding, in which the generator is trying to generate vertex pairs, while the discriminator tries to distinguish the generated vertex pairs from real connections (edges) in the network. Wasserstein-1 distance is adopted to train the generator to gain better stability. We develop three variations of models, including GANE which applies cosine similarity, GANE-O1 which preserves the first-order proximity, and GANE-O2 which tries to preserves the second-order proximity of the network in the low-dimensional embedded vector space. We later prove that GANE-O2 has the same objective function as GANE-O1 when negative sampling is applied to simplify the training process in GANE-O2. Experiments with real-world network datasets demonstrate that our models constantly outperform state-of-the-art solutions with significant improvements on precision in link prediction, as well as on visualizations and accuracy in clustering tasks.


\end{abstract}

%
%

\keywords{Network Embedding, Generative Adversarial Model, Wasserstein distance, Link Prediction}

\maketitle

\section{Introduction}

Representation learning, which provides low-dimensional vector-space representations for the data, is an important research field in machine learning since it can significantly simplify the algorithms. Representation learning for networks, a.k.a. network embedding, is specifically important for applications with massive amount of network-style of data, such as social networks and email graphs. The purpose of network embedding is to generate informative numerical representations of nodes and edges, which in turn enable further inference in the network data, such as link prediction and visualization. 

Most existing methods in network embedding explicitly define the representative structures of the network as some numerical/computational measurements during the representation learning process. For example, LINE \cite{LINE} and SDNE\cite{SDNEkdd16} use local-structure (first-order proximity) and/or global structure (second-order proximity), while DeepWalk \cite{deepwalk} uses network community structure. The learned representations are then sent to some machine learning toolkits to guide a specific discriminative task, such as link prediction with classification or regression models. That is, the network embedding is learned separately from the actual tasks. Therefore, the learned representations may not be capable of optimizing the objective functions of tasks directly. 

Alternative solutions utilize deep models to retrieve low-dimens-\\ional network representations. For example, Li et al. \cite{ijcaiLiWYL17} use a variational autoencoder to represent an information network. Thus, the representations obtained right before the decoder layer are considered as a learned representations when the reconstruction loss converges. HNE\cite{HNE2015kdd} studies network embedding for heterogeneous data. It integrates deep models into a unified framework to solve the similarity prediction problem. Similarly, the output of the layer before the prediction layer in HNE is treated as the embeddings. In these approaches, the network embedding is a by-product of applying deep models on a specific task.

However, the networking embeddings in existing solutions are somewhat handcrafted structures. Moreover, there is no systematical support to exhaustively explore potential structures of networks. Therefore, this paper proposes to incorporate Generative adversarial networks (GANs) into network embeddings.

GANs\cite{GAN} are promising frameworks for various learning tasks, especially in computer vision area. Technically, a GAN consists of two components: a discriminator trying to predict whether a given data instance is generated by the generator or not, and a generator trying to fool the discriminator. Both components are trained by playing minimax games from game theory. Various extensions have been proposed to address some theoretical limitations in GANs. For example, \cite{infoGAN}\cite{seqGAN} introduced modifications to GAN objectives, WGAN \cite{wGAN} confined the distribution measurement to improve the training stability, and IRGAN \cite{irGAN} extended the application domains to information retrieval tasks.

Motivated by the empirical success of the adversarial training process, we propose Generative Adversarial Network Embedding (GANE) to perform the network embedding task. To simplify the discussion, GANE only considers single relational networks in which the edges are of the same type in comparison to multi-relational networks with various types of edges. Hereafter, network embedding stands for network embedding for single relational networks.

In GANE, the generator tries to generate potential edges for a vertex and construct the representations for these edges, while the discriminator tries to distinguish the generated edges from real ones in the network and construct its own representations. Besides using cosine similarity, we also adopted the first-order proximity to define the loss function for the discriminator, and to measure the structural information of the network preserved in the low-dimensional embedded vector space. Under the principles of minimax game, the generator tries to simulate the structures of the network with the hints from the discriminator, and the discriminator in turn exploits the underlying structure to recover missing links for the network. Wasserstein-1 distance is adopted to train the generator with improved stability as suggested in WGAN\cite{wGAN}. To the best of our knowledge, this is the first attempt to learn network embedding in a generative adversarial manner. Experiments on link prediction and clustering tasks were executed to evaluate the performance of GANE. The results show that network embeddings learned in GANE can significantly improve the performance for supervised discrimination tasks in comparison with existing solutions. The main contributions of this paper can be summarized as follows:

\begin{itemize}
	\item We develop a generative adversarial framework for network embedding. It is capable of performing the feature representation learning and link prediction simultaneously under the adversarial minimax game principles.
	\item We discuss three variations of models, including naive GANE which applies cosine similarity, GANE-O1 which preserves the first-order proximity, and GANE-O2 which tries to preserves the second-order proximity of the network in the low-dimensional embedded vector space. 
	\item We evaluate the proposed models with detailed experiments on link prediction and clustering tasks. Results demonstrate significant and robust improvements in comparison with other state-of-the-art network embedding approaches.
\end{itemize}

The rest of the paper is organized as follows. Section 2 summarizes the related work. Section 3 illustrates the design and algorithms in GANEs. Section 4 presents the experimental design and results. Section 5 concludes the paper.


\section{Related Work}


The paper roots into two research fields, network embedding and generative adversarial networks.

\subsection{Network Embedding}

Extensive efforts have been devoted into the research about network embedding in recent years. Graph Factorization \cite{GF} represents the network as an affinity matrix of graph, and then utilizes a distributed matrix factorization algorithm to find the low-dimensional representations of the graph. DeepWalk \cite{deepwalk} utilizes the distribution of node degree to model topological structures of the network via the random walk and skip-gram to infer the latent representations of vertices in networks. Tang et al. proposed LINE \cite{LINE} to preserve both local (first-order) structures and global (second-order) structures during the embedding process by minimizing the KL-divergence between the distributions of structures in the original network and the embedded space. LINE has been considered as one of the most popular network embedding approaches in the past two years. Thereafter, Wang et al. proposed modularized nonnegative matrix factorization to incorporate the community structures and preserve such structures during representation learning \cite{CPNEaaai2017}. SDNE \cite{SDNEkdd16} applies a semi-supervised deep learning framework for network embedding, in which the first-order proximity is preserved by penalizing the similar vertices but far away in the embedded space and the second-order proximity is preserved by using a deep autoencoder. Li et al. \cite{content2vector} incorporated the text information and structure of the network into the embedding representations by employing the variational autoencoder(VAE) \cite{VAE}. Chang et al. proposed HNE\cite{HNE2015kdd} to address network embedding tasks with heterogeneous information, in which deep models for content feature learning and structural feature learning are integrated in a unified framework.

In summary, existing solutions in networking embedding either use shallow models or deep models to preserve different structural properties in the low-dimensional space, such as the connectivity between the vertices (the first order proximity), neighborhood connectivity patterns (the second order proximity), and other high-order proximities (the community structure). However, these solutions employ handcrafted structures for the network embedding, and it is hard to exhaustively explore potential structures of networks due to the lack of systematical support. Therefore, we propose to leverage on generative adversarial framework to explore potential structures in the networks to achieve more informative representations.


\subsection{Generative Adversarial Networks}
Recent advances in Generative Adversarial Networks (GANs)\cite{GAN} have proven GANs as a powerful framework for learning complex data distributions. The core idea is to define the generator and the discriminator to be the minimax game players competing with each other to push the generator to produce high quality data to fool the discriminator. 

Mirza \& Osindero introduced conditional GANs \cite{conditionalGAN} to control the data generation by setting conditional constraints on the model. InfoGAN \cite{infoGAN}, another information-theoretic extension to the GAN model, maximizes the mutual information between a small subset of the latent variables and the observations to learn interpretable and meaningful hidden representations on image datasets. SeqGAN\cite{seqGAN} models the data generator as a stochastic policy in reinforcement learning and uses the policy gradient to guide the learning process bypassing the generator differentiation problem for discrete data output.


Despite their successes, GANs are notably difficulty to train and prone to mode collapse \cite{arjovsky2017towards}, especially for discrete data. Energy-based GAN (EBGAN) \cite{ebGAN} tries to achieve a stable training process by viewing the discriminator as an energy function that attributes low energies to the regions near the data manifold and higher energies to other regions. However, EBGANs, which regularize the distribution distance as Jensen-Shannon (JS) divergence, share the same problem as classical GANs that the discriminator cannot be trained well enough, as the distance EBGANS adopted cannot offer perfect gradients. Replacing JS with the Earth Mover (EM) distance, Wasserstein-GAN \cite{wGAN} theoretically and experimentally solves the problem of model fragility.


GANs are successfully applied in the field of computer vision for tasks including generating sample images. However, there are few attempts to apply GANs to other machine learning tasks. Recently, IRGAN \cite{irGAN} has been proposed as an information retrieval model in which the generator focuses on predicting relevant documents given a query and the discriminator focuses on distinguish whether the generated documents are relevant. It showed superior performance over the state-of-the-art information retrieval approaches.




In this paper, we propose to explore the strength of generative adversarial models for network embedding. The proposed framework, GANE, performs the feature representation learning and link prediction simultaneously under the adversarial minimax game principles. Wasserstein-1 distance is adopted to define the overall objective function \cite{wGAN} to overcome the notorious unstable training problem in conventional GANs.

\section{GANE: Generative Adversarial Network Embedding}
\label{sec:GANE}


A network $N$ can be modeled as a set of vertices $V$ and a set of edges $E$. That is, a network can be represented as $N = (V, E)$. The primal task of network embedding is to learn a low-dimensional vector-space representation for each vertex $v_i \in V$. Unlike existing approaches which need to train the embedding presentation before applying it to predictive tasks, we facilitate predictions and the embedding learning process in a unified framework by leveraging on generative adversarial model.

\subsection{Naive GANE Discriminator}
\label{sec:naive_GANE}

\begin{figure}[t]
\flushleft
\includegraphics[width=0.5\textwidth]{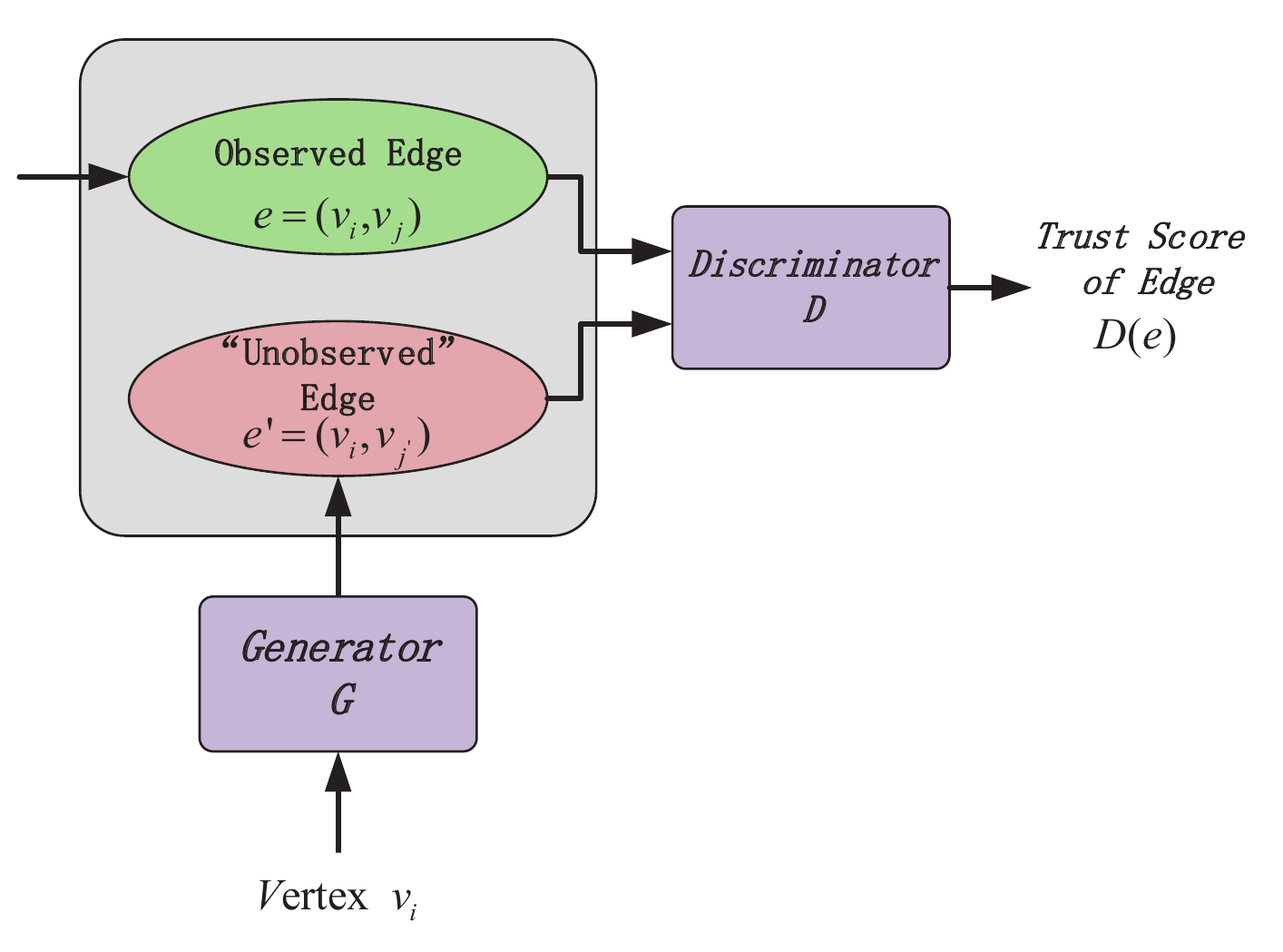}
\caption{Architecture and dataflow in GANE.}
\label{fig:gane}
\end{figure}

Two components, the generator $G$ and the discriminator $D$, are defined in GANE to play the minimax game. The task of $G$ is to predict the well-matched edge $(v_i, v_j)$ given $v_i$, while the task of $D$ is to identify the observed edges from the "unobserved" edges, where the "unobserved" edges are generated by $G$. The overall architecture and dataflow of GANE are depicted in Fig.\ref{fig:gane}.

To avoid the problem of unstable training in conventional GAN models which are prone to mode collapse, the Earth-Mover (also called Wasserstein) distance $W(\mathcal P_{E'}, \mathcal P_E)$ is utilized to define the overall objective function as suggested in WGAN \cite{wGAN}. $W(\mathcal P_{E'}, \mathcal P_E)$ can be informally defined as the minimum cost transporting mass from the distribution $\mathcal P_{E'}$ into the distribution $\mathcal P_E$. With mild assumptions, $W(\mathcal P_{E'}, \mathcal P_E)$ is continuous and differentiable almost everywhere. Following WGAN, the objective function is defined after the Kantorovich-Rubinstein duality\cite{Villani2009Optimal}:

\begin{equation}
\min_{G_\theta} \max_{D_\phi} ( \mathbb E_{e\sim \mathcal P_E} [D(e)]- \mathbb E_{e'\sim \mathcal P_{E'}} [D(e')] )
\end{equation}
\emph{where $\mathcal P_E$ is the distribution of observed edges, and $\mathcal P_{E'}$ is the distribution of unobserved edges generated by $G$}. That is, $e'=(v_i,v_j) \in G(v_i)$ and $G(v_i) \sim p_\theta(v_j|v_i)$. $D(e)$ is the utility function which computes the trust score (a scalar) for a given edge $e = (v_i, v_j)$. 

A naive version of GANE, or GANE, is defined with the following scoring function without considering the structural information in the network:
\begin{equation}
D(e) = s(v_i,v_j) = cos(\vec u_i, \vec u_j)=\frac{\vec u_i^T \vec u_j}{|\vec u_i|\cdot |\vec u_j|}
\label{fun:cosine_score}
\end{equation}
where $\vec u_i, \vec u_j \in \mathcal{R}^d$ are the embedding representation of vertex $v_i$ and $v_j$ respectively.

The discriminator is trained to assign high trust score to an observed edge $e$ but lower score to an unobserved edge $e'$ generated by G, while the generator is trained to produce contrastive edges with maximal trust score. Theoretically, there is a Nash equilibrium in which the generator perfectly fits in with the distribution of observed edges in the network (i.e., $ \mathcal P_{E'}=\mathcal P_E$), and the discriminator cannot distinguish true observed edges from the generated ones. However, it is computationally infeasible to reach such an equilibrium because the distribution of embedding in the low-dimensional space keeps changing dynamically along with the model training process. Consequently, the generator tends to learn the distribution $\mathcal P_{E'}$ to model the network structure as accurately as possible, while the discriminator tends to accept the potential true (unobserved but in all probability be true) edges. 



\subsection{Structure-preserving Discriminator}
\label{sec:GANE_O1}

Structural information of the network may provide valuable guidance in the model learning process. Therefore, we propose to extend the discriminator definition in GANE with the concepts of first-order and second-order proximity, which were introduced in LINE\cite{LINE}.

\begin{definition}
(\textbf{First-order Proximity}) \emph{The first-order proximity in a network describes the local pairwise proximity between two vertices. The strength between two vertices $v_i$ and $v_j$ is denoted as $w_{ij}$. $w_{ij}=0$ indicates there is no edge observed between $v_i$ and $v_j$}.
\end{definition}

The intuitive solution is to embed the vertices with strong ties (i.e. high $w_{ij}$) close to each other in the low-dimensional space. Therefore, $w_{ij}$ can be used as the weighting factor to evaluate the embedding representation.

For network embedding, the goal is to minimize the difference between the probability distribution of the edges in the original space and that in the embedded space. The distance between the empirical probability distribution $\hat p_1(\cdot, \cdot)$ and the resulting probability distribution $p_1(\cdot, \cdot)$ in the network embedding can be defined as

\begin{equation}
\begin{aligned}
O_1 &= distance(\hat p_1(\cdot, \cdot), p_1(\cdot, \cdot)) \\
&= -\sum_{(v_i,v_j)\in E} w_{ij}\log p_1(v_i,v_j)
\end{aligned}
\end{equation}
where $p_1(v_i, v_j)$ is the joint probability between vertex $v_i$ and $v_j$ and $E$ is the set of observed edges in the network. The empirical probability is defined as $\hat{p_1}(i,j)=w_{ij}/W$, where $W=\sum_{(v_i,v_j)\in E} w_{ij}$. For each edge $(v_i, v_j)$, $p_1(v_i, v_j)$ is defined as 
\begin{equation}
p_1(v_i, v_j) = \sigma(\vec u_i^T \vec u_j)
\label{fun:p1}
\end{equation}
Following Eq.(1), it is equivalent for the discriminator to minimize the loss of GANE-O1, which is GANE with first-order proximity, as 
\begin{equation}
\begin{aligned}
\mathcal L_{D_\phi}^{O_1} &=& \mathbb E_{e'\sim \mathcal P_{E'}} [D(e')]-\mathbb E_{e\sim \mathcal P_E} [D(e)] \\
&=& \mathbb E_{(v_i,v_{j'})\sim \mathcal P_{E'}} [w_{ij'}\log p_1(v_i,v_{j'})] \\
&&-\mathbb E_{(v_i,v_j)\sim \mathcal P_E} [w_{ij}\log p_1(v_i,v_j)] \\
&=& \mathbb E_{(v_i,v_{j'})\sim \mathcal P_{E'}} [w_{ij'}\log \sigma(\vec u_i^T \vec u_{j'})] \\
&&-\mathbb E_{(v_i,v_j)\sim \mathcal P_E} [w_{ij}\log \sigma(\vec u_i^T \vec u_j)].
\end{aligned}
\label{fun:GANE_O1}
\end{equation}

\begin{definition}
(\textbf{Second-order Proximity}) \emph{The second-order proximity between a pair of vertices $(v_i,v_j)$ describes the similarity between their neighborhood structure in the network. Let $\vec W_i=(w_{i1},w_{i2},\\...,w_{i|V|})$ denote the first-order proximity of $v_i$ with other vertices. Then, the second-order proximity between $v_i$ and $v_j$ is determined by the similarity between $\vec W_i$ and $\vec W_j$.}
\end{definition}

The intuitive solution is to embed the vertices which have high second-order proximity close to each other in the low-dimensional space. By analogy with the corpus in natural language processing (NLP), the neighbors of $v_i$ can be treated as its context (nearby words), and a vertex $v_j$ with similar context is considered to be similar. Similar to the skip-gram model \cite{skip_gram}, the probability of "context" $v_j$ generated by $v_i$ is defined with the softmax function as
\begin{equation}
p_2(v_j|v_i)=\frac{exp(\vec u_j^T \vec u_i)}{\sum_{k=1}^{|V|}exp(\vec u_k^T \vec u_i)}.
\end{equation}
\begin{algorithm}[t]
\SetAlgoNoLine
\KwIn{$\alpha$, the learning rate. $c$, the clipping parameter. $m$, the batch size. $M$, the number of generated edges given a source vertex. $\mathcal T$, the training number for $D$.}
\KwIn{$V$, the set of vertices in the network. $E$, the set of edges in the network.}

Randomly initialize $\phi$, $\theta$

\Repeat{convergence}{
        \For{$t=0,...,\mathcal T-1$
    }{
      Sample $\{e_k\}^m_{k=1} \sim \mathcal P_E$, a batch of edges from the network.

      Sample $\{e'_k\}^m_{k=1} \sim \mathcal P_{E'}$, a batch of edges from the generated pool.
      
      Update $\phi$ to minimize $\frac{1}{m}\sum_{k=1}^m D(e_k')-\frac{1}{m}\sum_{k=1}^m D(e_k)$

      $clip(\phi,-c,c)$
        }

      Sample $\{v_i\}^m_{i=1} \sim V$, a batch of vertices from the network.

      \For{each $v_i$
      }{
      Sample $\{e_j=(v_i,v_j)\}^M_{j=1} \sim \mathcal P_{E'}$
      }

      Update $\theta$ to minimize $-\frac{1}{m}\sum_{i=1}^m(\frac{1}{M}\sum_{j=1}^M D(e_j))$
      }
\caption{Network Embedding Learning in GANE}
\label{alg:GANE}
\end{algorithm}
The objective function, which is the distance between the empirical conditional distribution $\hat p_2(\cdot|v_i)$ and the resulting conditional distribution in the embedded space $p_2(\cdot|v_i)$, is defined as
\begin{equation}
\begin{aligned}
O_2 &= \sum_{v_i\in V}\lambda_{v_i} distance(\hat p_2(\cdot|v_i), p_2(\cdot|v_i)) \\
&= -\sum_{v_i\in V}\sum_{\{j|(v_i,v_j)\in E\}} \lambda_{v_i} \hat p_2(v_j|v_i) \log p_2(v_j|v_i)
\end{aligned}
\label{fun:LINE_O2}
\end{equation}
where $\lambda_{v_i}$ denotes the prestige of $v_i$ in the network. For simplicity, the sum of weights in $\vec W_i$ is used as the prestige of $v_i$. That is, $\lambda_{v_i}=\sum_{j=1}^{|V|}w_{ij}$. The empirical distribution $\hat p_2(\cdot|v_i)$ is defined as
\begin{equation}
\hat p_2(v_j|v_i)=\frac{w_{ij}}{\sum_{j=1}^{|V|}w_{ij}}.
\end{equation}
Then, Eq.(\ref{fun:LINE_O2}) can be rewritten as
\begin{equation}
\begin{aligned}
O_2 &=-\sum_{(v_i,v_j)\in E}w_{ij}\log p_2(v_j|v_i)\\
&=-\sum_{(v_i,v_j)\in E}w_{ij}\log (\frac{exp(\vec u_j^T \vec u_i)}{\sum_{k=1}^{|V|}exp(\vec u_k^T \vec u_i)}).
\end{aligned}
\label{fun:LINE_O2_simplier}
\end{equation}
However, the computation of the objective function Eq.(\ref{fun:LINE_O2_simplier}) remains expensive because the softmax term $p_2(v_j|v_i)$ needs to sum up all vertices of the network. A general solution is to apply negative sampling \cite{DBLP:journals/corr/Dyer14} to bypass the summations. This solution is based on Noise Contrastive Estimation (NCE) \cite{NCE} which shows that a good model should be able to differentiate data from noise by means of logistic regression. With the method of negative sampling,  
\begin{equation}
\log p_2(v_j|v_i) = \log \sigma(\vec u_j^T \vec u_i)+\sum_{k=1}^{K}\mathbb E_{v_k\sim\mathcal P_k(v)}[\log\sigma(-\vec u_k^T \vec u_i)].
\label{fun:LINE_O2_negative_sampling}
\end{equation}

By replacing $D(e)$ in Eq.(1) with updated Eq.(\ref{fun:LINE_O2_negative_sampling}) via aforementioned negative sampling counterpart, it is equivalent for the discriminator to minimize the loss of GANE-O2, which is GANE with second-order proximity, as
\begin{equation}
\begin{split}{}
&\mathcal L_{D_\phi}^{O_2} \\
&= \mathbb E_{e'\sim \mathcal P_{E'}} [D(e')]-\mathbb E_{e\sim \mathcal P_E} [D(e)] \\
&=\mathbb E_{(v_i,v_{j'})\sim \mathcal P_{E'}} [w_{ij'}(\log \sigma(\vec u_{j'}^T \vec u_i)+\sum_{k'=1}^{K}\mathbb E_{v_{k'}\sim\mathcal P_k(v)}[\log\sigma(-\vec u_{k'}^T \vec u_i)])] \\
&-\mathbb E_{(v_i,v_j)\sim \mathcal P_E} [w_{ij}(\log \sigma(\vec u_j^T \vec u_i)+\sum_{k=1}^{K}\mathbb E_{v_k\sim\mathcal P_k(v)}[\log\sigma(-\vec u_k^T \vec u_i)])]. \\
\end{split}
\label{fun:GANE_O2}
\end{equation}

\begin{table*}[t]
  \caption{The Optimization Direction and Ranking Score for Comparison}
  \label{tab:criteria_score}
  \begin{tabular}{lllll}
    \toprule
    \multirow{2}{0.12\textwidth}{Models}& Distribution & \multirow{2}{0.15\textwidth}{Objective Function}&\multirow{2}{0.15\textwidth}{Optimization Direction}&Ranking \\
    &Measurement & &&Score\\
    \midrule
    LINE-O1& KL-divergence& $\min \sum_{(v_i,v_j)\in E}-w_{ij}\log p_1(v_i,v_j)$& $p_1(v_i,v_j)=\sigma(u_i^T u_j)$& $u_i^T u_j$\\
    \midrule
    LINE-O2& KL-divergence&$\min \sum_{(v_i,v_j)\in E}-w_{ij}\log p_2(v_j|v_i)$& $p_2(v_j|v_i)=\frac{exp(u_i^T u_j)}{\sum_{k=1}^{|V|}exp(u_i^T u_k)}$& $u_i^T u_j$\\

    \midrule
    \multirow{2}{0.12\textwidth}{LINE-(O1+O2)}
    &\multirow{2}{0.12\textwidth}{KL-divergence}
    &$\min \sum_{(v_i,v_j)\in E}-w_{ij}\log p_1(v_i,v_j)$
    &$p_1(v_i,v_j)=\sigma(u_i^T u_j)$
    & \multirow{2}{0.08\textwidth}{$u_i^T u_j$}\\
    &&$\min \sum_{(v_i,v_j)\in E}-w_{ij}\log p_2(v_j|v_i)$& $p_2(v_j|v_i)=\frac{exp(u_i^T u_j)}{\sum_{k=1}^{|V|}exp(u_i^T u_k)}$&\\

    \midrule
    \multirow{2}{0.12\textwidth}{IRGAN}
    & \multirow{2}{0.15\textwidth}{JS-divergence}
    & {$\min_\theta\max_\phi \sum(\mathbb E_{e=(v_i,v_j)\sim \mathcal P_E[\log D(e)]}$}
    & \multirow{2}{0.15\textwidth}{$D(e)=\sigma (u_i^T u_j)$}
    & \multirow{2}{0.08\textwidth}{$u_i^T u_j$}\\
    &&$ + \mathbb E_{e'=(v_i,v_{j'})\sim\mathcal P_{E'}}[log(1-D(e'))]) $&&\\

    \midrule
    \multirow{2}{0.12\textwidth}{GANE}
    & \multirow{2}{0.15\textwidth}{Wasserstein distance}
    & {$\min_{\theta} \max_{\phi} ( \mathbb E_{e=(v_i,v_j)\sim \mathcal P_E} [D(e)]$}
    &\multirow{2}{0.17\textwidth}{$D(e)=cosine(u_i,u_j)$}
    & \multirow{2}{0.08\textwidth}{$cosine(u_i,u_j)$}\\
    &&$- \mathbb E_{e'=(v_i,v_{j'})\sim \mathcal P_{E'}} [D(e')] )$&&\\
    \midrule
    \multirow{2}{0.12\textwidth}{GANE-O1}
    & \multirow{2}{0.15\textwidth}{Wasserstein distance}
    &{$\min_{\theta} \max_{\phi}(\mathbb E_{e=(v_i,v_j)\sim \mathcal P_E} [w_{ij}\log p_1(v_i,v_j)]$}
    & \multirow{2}{0.15\textwidth}{$p_1(v_i,v_j)=\sigma(u_i^T u_j)$}
    & \multirow{2}{0.08\textwidth}{$u_i^T u_j$}\\
    &&$- E_{e'=(v_i,v_{j'})\sim \mathcal P_{E'}} [w_{ij'}\log p_1(v_i,v_{j'})])$&&\\
  \bottomrule
\end{tabular}
\end{table*}


The noise distribution $\mathcal P_k(v)$ is empirically set to $3/4$ power \cite{skip_gram}\cite{LINE}. That is, $\mathcal P_k(v) \varpropto W_v^{3/4}$, where $W_v = \sum_{j=1}^{|V|}w_{vj}$. In general, the larger number of negative sampling $K$, the better performance of the model. Moreover, $\sum_{k'=1}^{K}\mathbb E_{v_{k'}\sim\mathcal P_k(v)}[\log\sigma(-\vec u_{k'}^T \vec u_i)]$ and $\sum_{k=1}^{K}\mathbb E_{v_k\sim\mathcal P_k(v)}[\log\sigma(-\vec u_k^T \vec u_i)]$ will be equivalent when $K$ is infinite. Therefore, Eq.(\ref{fun:GANE_O2}) can be updated as
\begin{equation}
\begin{aligned}
\mathcal L_{D_\phi}^{O_2} =& \mathbb E_{(v_i,v_{j'})\sim \mathcal P_{E'}} [w_{ij'}\log \sigma(\vec u_{j'}^T \vec u_i)]\\
&-\mathbb E_{(v_i,v_j)\sim \mathcal P_E} [w_{ij}\log \sigma(\vec u_j^T \vec u_i)]\\
&=\mathcal L_{D_\phi}^{O_1}
\end{aligned}
\label{fun:GANE_O2_opt}
\end{equation}
which shows that GANE-O1 (Eq.(\ref{fun:GANE_O1})) and GANE-O2 have the same objective function. For this reason, the rest of paper will only experiment and discuss GANE and GANE-O1.



\subsection{Generator Optimization}
In minimax game, the generator plays as an adversary of the discriminator, and it needs to minimize the loss function defined as (referring to Eq.(1)):
\begin{equation}
\mathcal L_{G_\theta} = - \mathbb E_{e'\sim \mathcal P_{E'}} [D(e')]
\label{fun:GANE_G}
\end{equation}

The generator of GANE is in charge of generating unobserved edges. Different from sampling random variables during generating process in conventional GANs\cite{GAN,conditionalGAN}, GANE requires $v_j$ to be a real vertex in the network when it generates/predicts an unobserved edge $(v_i,v_j)$ for a given $v_i$. As the sampling of vertex $v_j$ is discrete, Eq.(\ref{fun:GANE_G}) cannot be optimized directly. Inspired by SeqGAN \cite{seqGAN}, the policy gradient which is frequently used in reinforcement learning\cite{gradientForRL} is applied. The derivation of the policy gradient for GANE generator $G$ is computed as
\begin{equation}
\begin{aligned}
\nabla_\theta \mathcal L_{G_\theta} &= \nabla_\theta (- \mathbb E_{e'\sim \mathcal P_{E'}} [D(e')])\\
&=-\sum_{n=1}^{N} \nabla_\theta \mathcal P_\theta(e_n')D_\phi(e_n')\\
&=-\sum_{n=1}^{N} \mathcal P_\theta(e_n')\nabla_\theta \log \mathcal P_\theta(e_n')D_\phi(e_n')\\
&=-\mathbb E_{e'\sim \mathcal P_{E'}}[\nabla_\theta \log \mathcal P_\theta(e')D_\phi(e')]\\
&\simeq -\frac{1}{M}\sum_{j=1}^{M}\nabla_\theta \log \mathcal P_\theta(e_j')D_\phi(e_j')
\end{aligned}
\end{equation}
where a sampling approximation is used in the last step. \emph{$e_j=(v_i,v_j)$ is a sample edge starting from a given source vertex $v_i$ following $\mathcal P_{E'}$, which is the distribution of the current version of generator}. The distribution $\mathcal P_{E'}$ is determined by the parameter $\theta$ of the generator. Every time $\theta$ is updated during the model training, a new version of distribution $\mathcal P_{E'}$ is generated. $M$ is the number of samples. 

In reinforcement learning terminology\cite{policyRL}, the discriminator acts as the environment for the generator, feeding a reward $D_\phi(e_j')$ to the generator $G$ when $G$ takes an action, such as generating/pred-\\icting an edge $(v_i,v_j)$ for a given $v_i$.
\begin{table*}[t]
  \caption{Binary Classification Performance Comparison For Link Prediction }
  \centering
  \label{tab:link_prediction_classification}
  \begin{tabular}{lccccccccc}
     \toprule
     Models&10\%&20\%&30\%&40\%&50\%&60\%&70\%&80\%&90\%\\
      \midrule
     LINE-O1 (\%)&73.12&75.77&77.51&78.39&79.18&79.40&79.92&80.09&80.33\\
     LINE-O2 (\%)&77.83&83.71&86.90&86.19&87.08&89.25&89.21&88.91&89.99\\
    LINE-(O1+O2) (\%)&82.18&86.73&85.03&89.40&91.74&90.65&92.32&92.44&93.06\\
    IRGAN (\%)&58.52&62.07&63.06&62.52&64.48&58.54&66.78&63.71&63.42\\
    GANE (\%)&\textbf{93.85}&\textbf{94.61}&\textbf{95.04}&\textbf{94.99}&\textbf{95.11}&\textbf{95.23}&\textbf{95.32}&\textbf{95.46}&\textbf{95.09}\\
    GANE-O1 (\%)&80.49&83.24&86.82&84.92&85.72&82.43&83.87&86.34&85.90\\
   \bottomrule
\end{tabular}
\end{table*}


\subsection{Model Training}
We randomly sample 90\% edges from the network as the training set for the training process, and enforce the requirement that these samples should cover all vertices. Therefore, the embedding of all vertices could be learned in our models. For each training iteration, the discriminator is trained for $\mathcal T$ times, but the generator is trained just once. Mini-batch Stochastic Gradient Descent and RMSProp\cite{RMSProp} optimizer based on the momentum method are applied as they perform well even on highly non-stationary problems. In order to have parameters $\phi$ lie in a compact space, the paper experiments with simple variants with little difference and sticks with parameters clipping. For more details, please refer to\cite{wGAN}. The overall algorithm for GANEs is provided in Algorithm \ref{alg:GANE}.

\begin{figure}[t]
\centering
\includegraphics[width=0.5\textwidth]{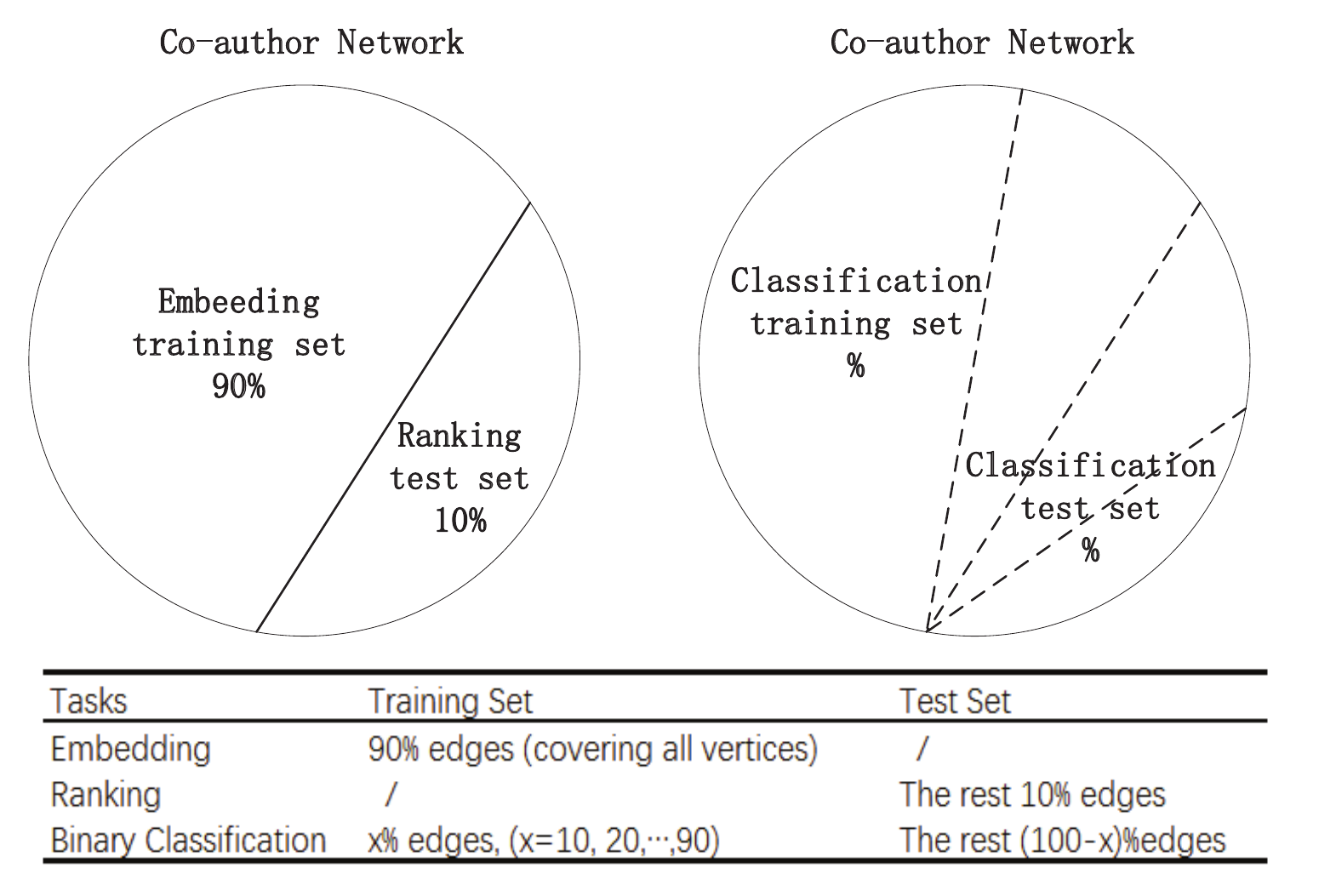}
\caption{Dataset Partition}
\label{fig:DatasetPartition}
\end{figure}

\section{Experimental Evaluation}
\label{sec:experiments}
To evaluate the proposed models, we applied the embedding representations to two task categories: link prediction, and clustering. For each category, we compared proposed GANE and GANE-O1 with several state-of-the-art approaches for network embedding. 

The list of models in comparison includes:
\begin{itemize}
\setlength{\itemsep}{5pt}


\item \textbf{LINE \cite{LINE}.} LINE is a very popular and state-of-the-art model for network embedding. Three variations of LINE were evaluated: LINE-O1, LINE-O2, LINE-(O1+O2). LINE-O1 and LINE-O2 consider only the first-order proximity and second-order proximity respectively. LINE-(O1+O2) utilizes the concatenated vectors from the outputs of LINE-O1 and LINE-O2.

\item \textbf{IRGAN \cite{irGAN}.} IRGAN was selected as a representative for GAN-related models. IRGAN is designed as a minimax game to improve the performance of information retrieval tasks. To enable comparison, we turned IRGAN into a model for network embedding by featuring parameters in IRGAN as the low-dimensional representations for the network.

\item \textbf{GANE.} The naive GANE model was defined in Section \ref{sec:naive_GANE}. It evaluates the trust score of an edge with the cosine distance between two vertices in low-dimensional vector space.

\item \textbf{GANE-O1.} The GANE model with the first-order proximity of the network was defined in Section \ref{sec:GANE_O1}. 
\end{itemize}


An overview about key technical definitions for these models is provided in Table~\ref{tab:criteria_score}.

The dataset used in the experiments is the co-author network constructed from the DBLP dataset\footnote{Available at http://arnetminer.org/citation} \cite{DBLP}. The co-author network records the number of papers co-published by authors. The co-author relation is considered as an undirected edge between two vertices (authors) in the network. Furthermore, the network we constructed consists of three different research fields: data mining, machine learning, and computer vision. It includes $10,541$ authors and $97,072$ edges. Each vertex (author) in the network is labeled according to the research areas of papers published by this author. The dimensionality of the embedding vectors is set to 128 for all models. 




\begin{table*}
  \caption{Ranking Performance Comparison for Link Prediction}
  \centering
  \label{tab:link_prediction_ranking}
  \begin{tabular}{lccccccccccc}
	\toprule
	Metric&P@1&P@3&P@5&P@10&MAP&R@1&R@3&R@5&R@10&R@15&R@20\\
	\midrule
	LINE-O1&0.0185&0.0378&0.0378&0.0326&0.0812&0.0120&0.0754&0.1230&0.2100&0.2694&0.3203 \\
	Improve&563.78\%&456.88\%&470.37\%&361.96\%&310.47\%&453.33\%&361.14\%&399.92\%&283.24\%&215.70\%&170.96\%\\
	\midrule
	LINE-O2&0.0&0.1124&0.1247&0.0921&0.2073&0.0&0.2483&0.4554&0.6409&0.6973&0.7278\\
	Improve&N/A&87.28\%&72.89\%&63.52\%&60.78\%&N/A&40.03\%&35.02\%&25.57\%&21.97\%&19.25\%\\
	\midrule
    LINE-(O1+O2)&0.0&0.0928&0.0905&0.0650&0.1535&0.0&0.1971&0.3128&0.4323&0.4917&0.5282\\
    Improve&N/A&126.83\%&138.23\%&131.69\%&117.13\%&N/A&76.41\%&96.58\%&86.17\%&72.97\%&64.31\%\\
    \midrule
    IRGAN&0.0231&0.1554&0.1665&0.1160&0.2681&0.0102&0.3311&0.5898&0.7750&0.8193&0.8406\\
    Improve&431.60\%&35.46\%&29.49\%&29.83\%&24.32\%&550.98\%&5.01\%&4.26\%&3.85\%&3.81\%&3.25\%\\
    \midrule
    GANE&0.0&0.1864&\textbf{0.2208}&\textbf{0.1598}&0.2978&0.0&0.3080&\textbf{0.6236}&\textbf{0.8459}&\textbf{0.8913}&\textbf{0.9112}\\
    Improve&N/A&12.93\%&-2.36\%&-5.76\%&11.92\%&N/A&12.89\%&-1.40\%&-4.86\%&-4.58\%&-4.75\%\\
    \midrule
    GANE-O1&\textbf{0.1228}&\textbf{0.2105}&0.2156&0.1506&\textbf{0.3333}&\textbf{0.0664}&\textbf{0.3477}&0.6149&0.8048&0.8505&0.8679\\
   \bottomrule
\end{tabular}
\end{table*}

\subsection{Link Prediction}
Link prediction tries to predict the missing neighbor $v_j$ in an unobserved edge $(v_i, v_j)$ for a given vertex $v_i$ of the network or to predict the likelihood of an association between two vertices. It is worth noting that the proposed models GANE and GANE-O1 both have implied answers for link prediction, as the generator is trained to produce the best answer of $v_j$ given $v_i$. Therefore, there is no need to train a binary classifier for link prediction, or to sort the candidates by a specific metric as most models usually do.

To make fair and impartial comparisons, we evaluated the link prediction task in two aspects as: 
\begin{enumerate}
	\item a binary classification problem by employing the embedding representations learned in models, and
    \item a ranking problem by scoring the pair of vertices which is represented as a low-dimensional vector.
\end{enumerate}

\subsubsection{Classification Evaluation}\label{sec:classification}
For binary classification evaluation, we used the Multilayer Perceptron (MLP) \cite{MLP} classifier to tell positive or negative samples. We randomly sampled different percentages of the edges as the training set and used the rest as the test set for the evaluation. 

In the training stage, the observed edges in the network were used as the positive samples, and the same size of unobserved edges were randomly sampled as negative samples. The embedding representations of two vertices of an edge were then concatenated as the input to the MLP classifier. 

\begin{figure}[thp]
\centering
\includegraphics[width=0.5\textwidth]{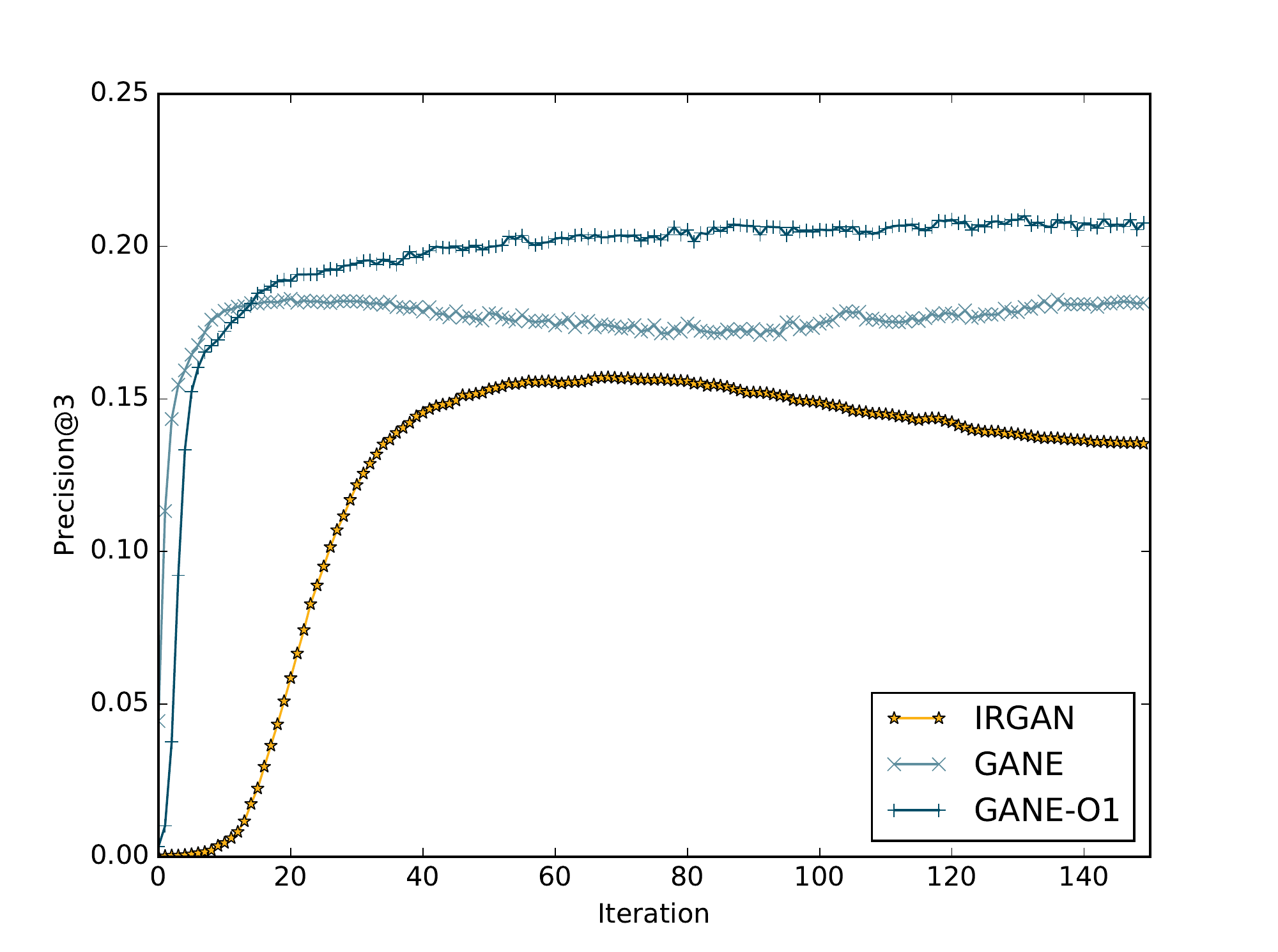}
\caption{Training Curves}
\label{fig:Training_Curves}
\end{figure}
In the test stage, records in the test set were fed into the classifier. 
Table~\ref{tab:link_prediction_classification} reports the accuracy of the binary classification achieved by different models. The results show that our models (both GANE and GANE-O1) outperform all the baselines consistently and significantly given different training-test cuttings. Moreover, they are quite robust/insensitive to the size of the training set in comparison with other approaches. Both GANE and GANE-O1 perform better than IRGAN which demonstrates the effectiveness of the adoption of Wasserstein-1 distance to GAN models. 

\begin{figure}[t]
\centering
\includegraphics[width=0.5\textwidth]{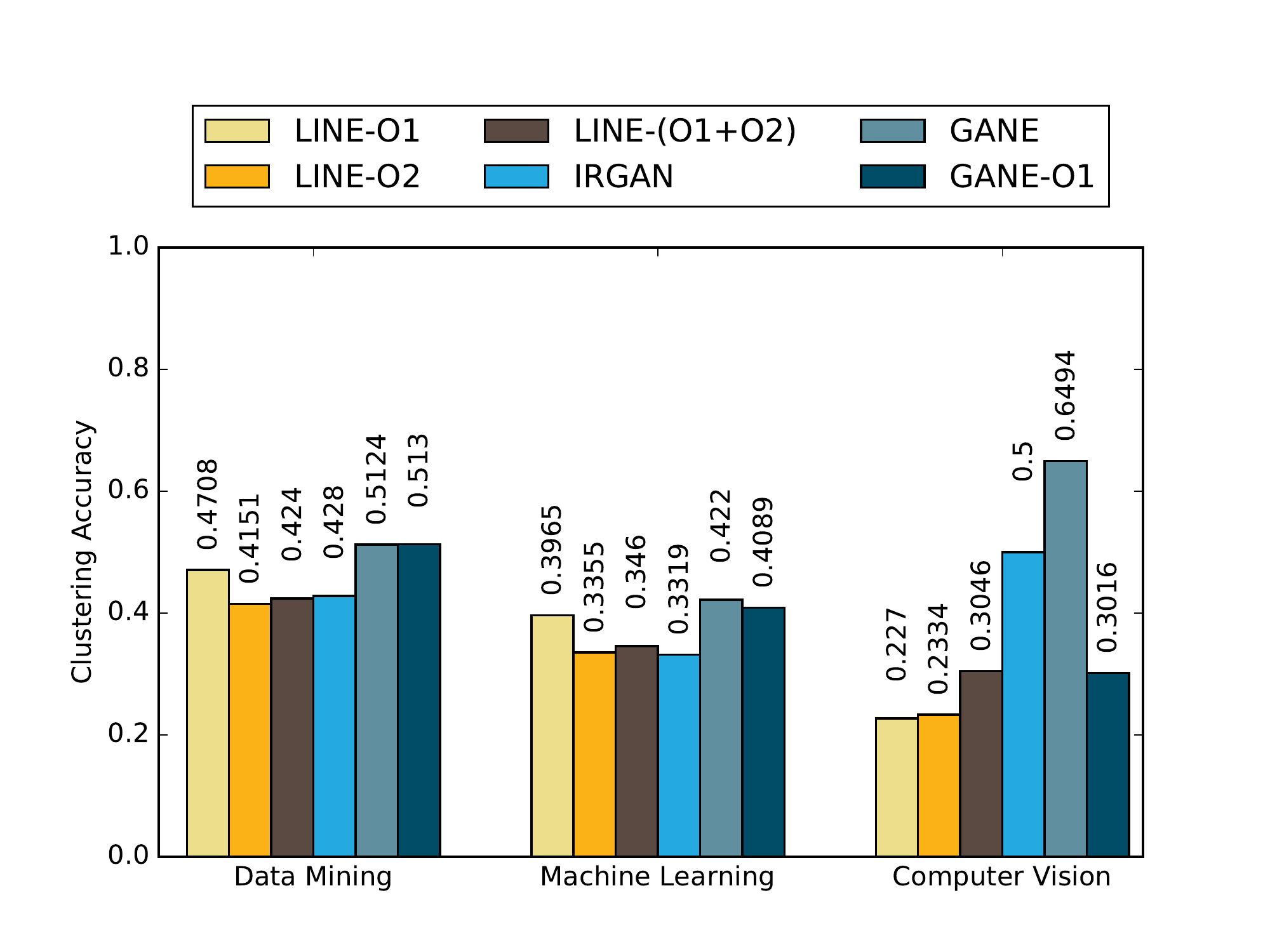}
\caption{Performance Comparison On Clustering.}
\label{fig:acc_category_Label_Classification}
\end{figure}

LINE-(O1+O2) has the best performance in the three variations of LINE, as it explores both first-order proximity and second order proximity which are the representative structures in the co-author network as suggested in \cite{LINE}. For the models explicitly adopt the first order proximity, GANE-O1 performs better than LINE-O1. Our guess is that the proposed generative adversarial framework for network embedding is capable of capturing and preserving more complex network structures implicitly. 

It is worth noting that GANE shows its full strength on the link prediction task even if it is simpler than GANE-O1 without considering the relationship between vertices in the network. This may be attributed to the fact that the co-author network is quite sparse.

\begin{figure*}[t]
\centering
\subfigure[GANE]{
	\includegraphics[width=0.32\textwidth]{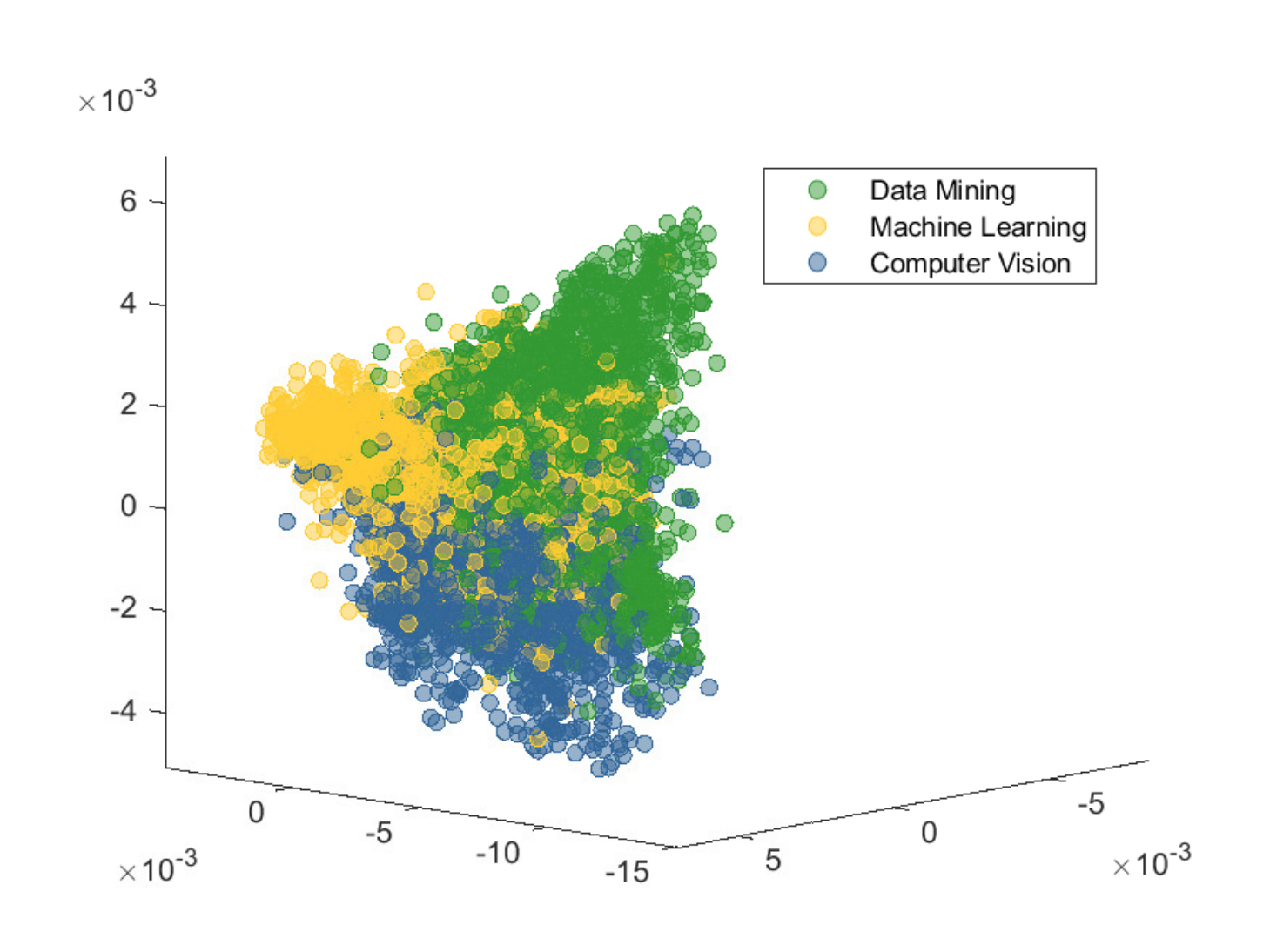}
	\label{fig:visulization_GANE}
}
\subfigure[GANE-O1]{
	\includegraphics[width=0.32\textwidth]{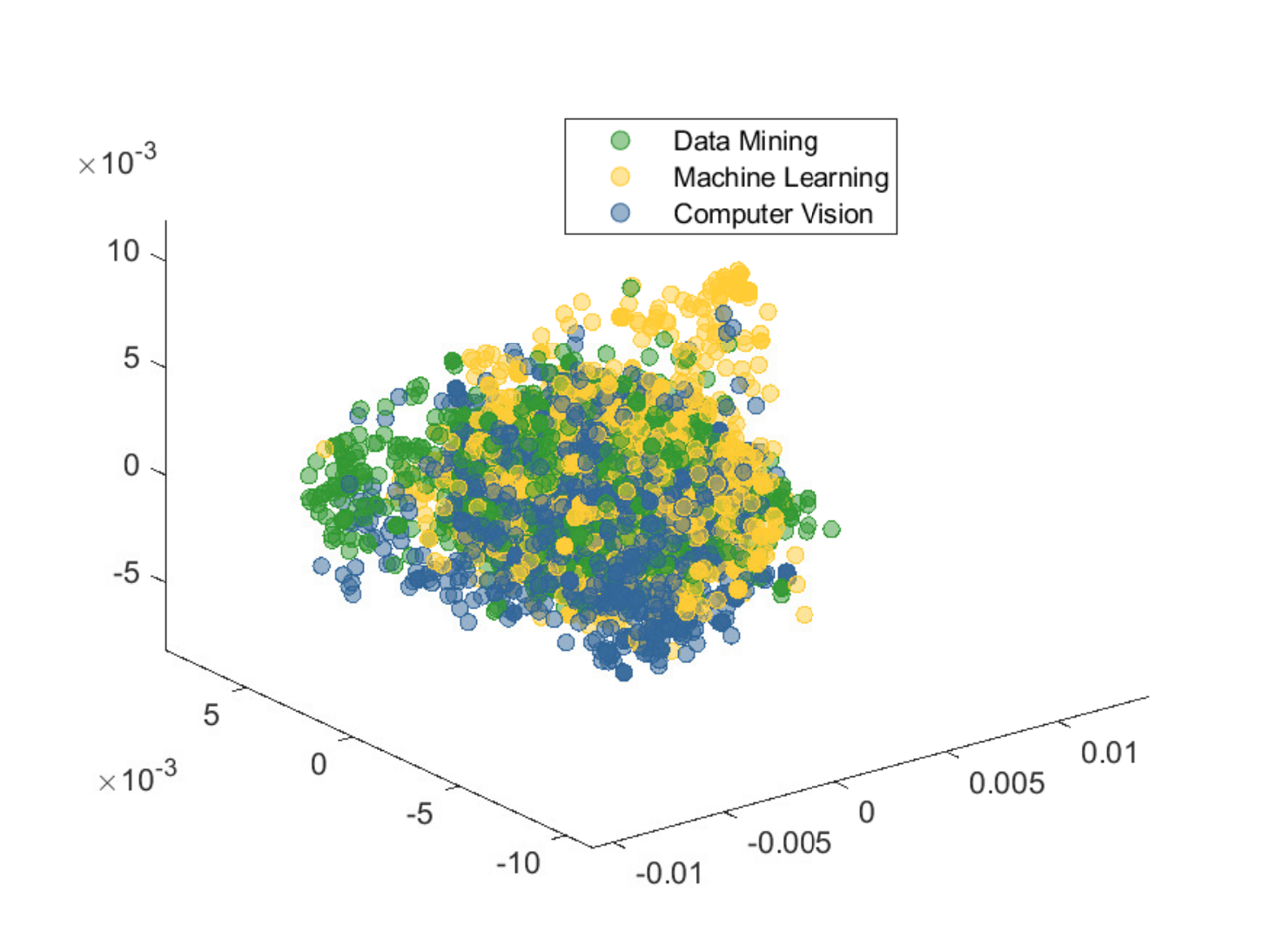}
	\label{fig:visulization_GANEo1}
}
\subfigure[IRGAN]{
	\includegraphics[width=0.32\textwidth]{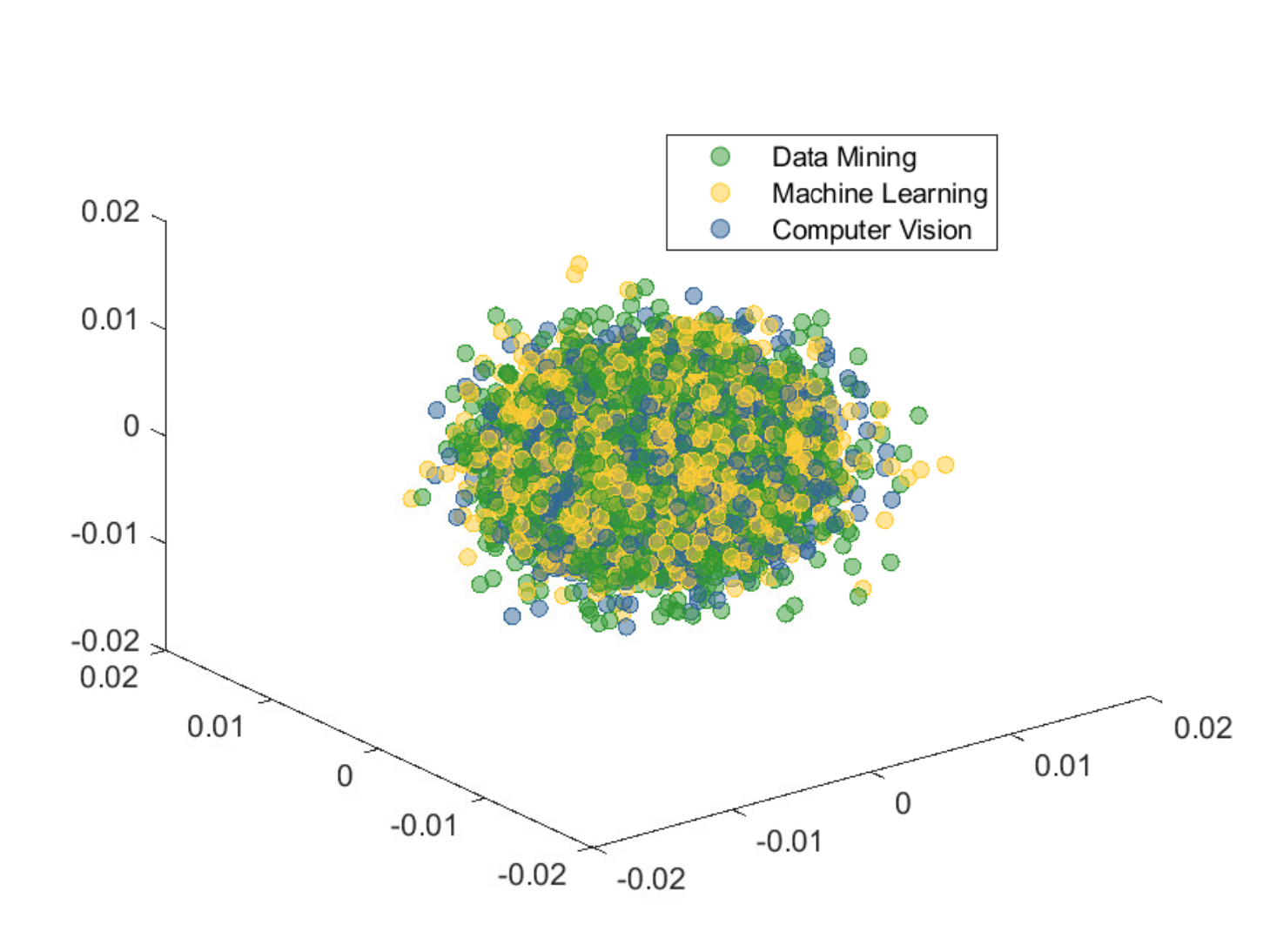}
	\label{fig:visulization_IRGAN}
}

\subfigure[LINE-O1]{
	\includegraphics[width=0.32\textwidth]{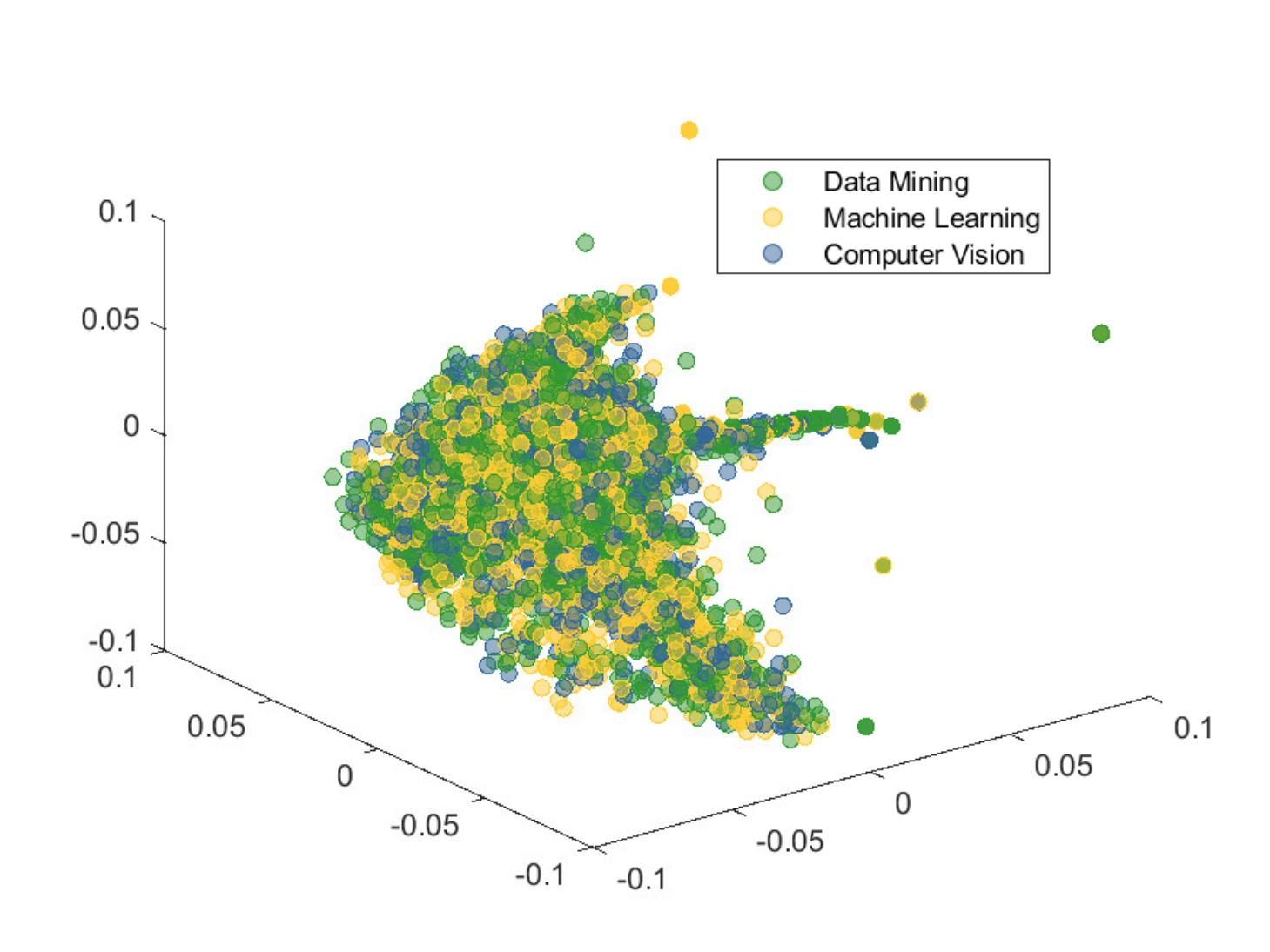}
	\label{fig:visulization_LINEo1}
}
\subfigure[LINE-O2]{
	\includegraphics[width=0.32\textwidth]{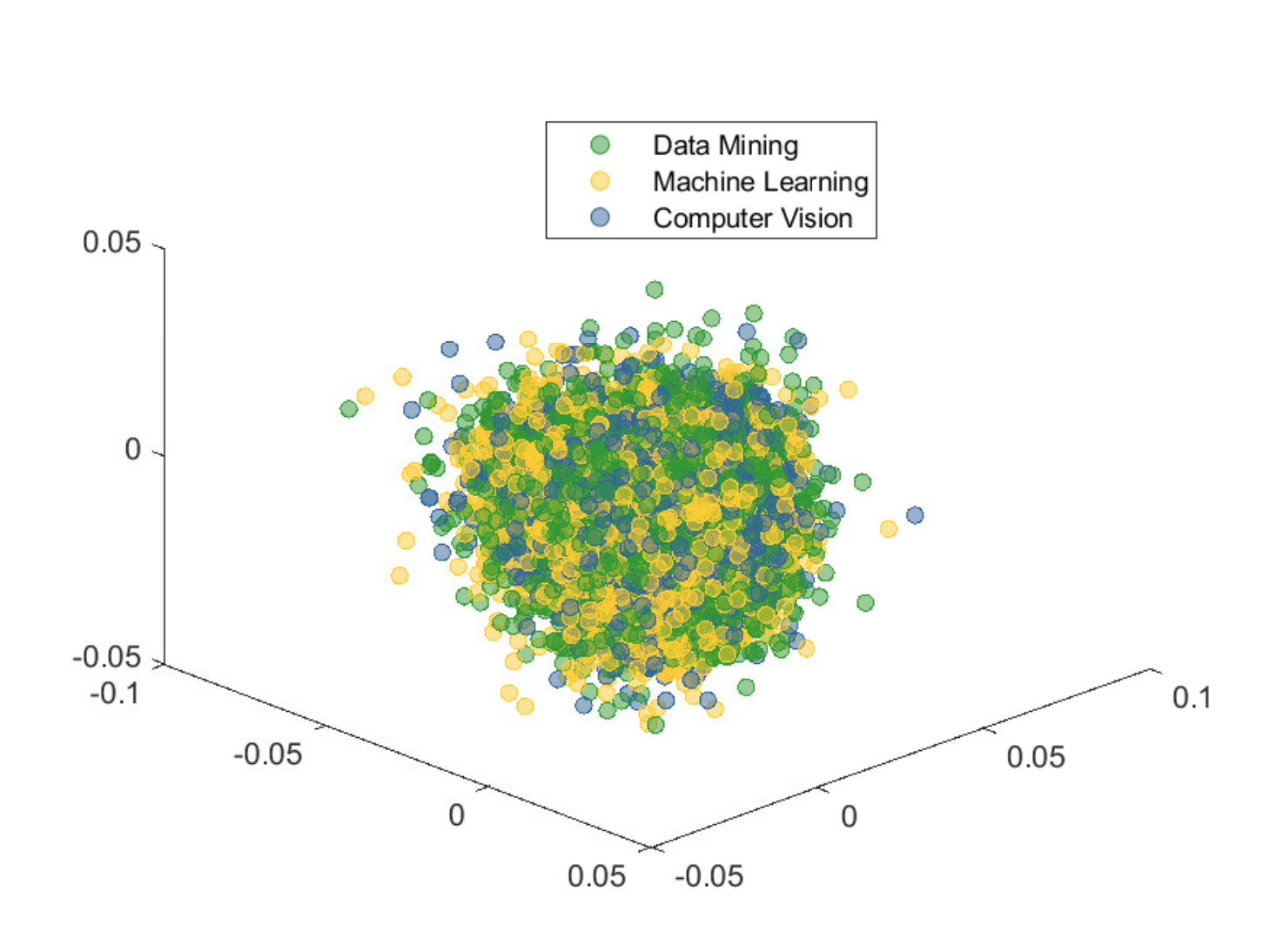}
	\label{fig:visulization_LINEo2}
}
\subfigure[LINE-(O1+O2)]{
	\includegraphics[width=0.32\textwidth]{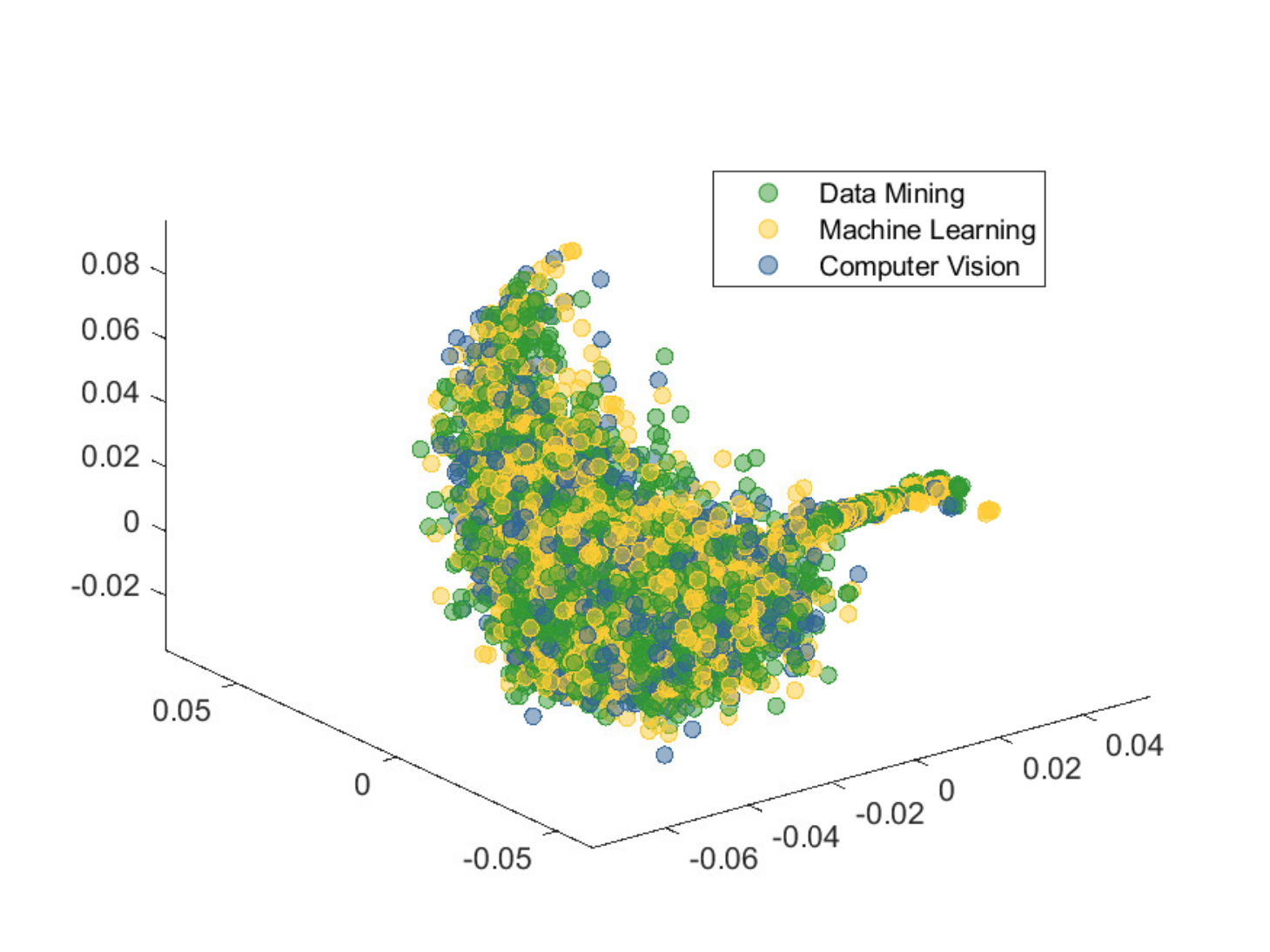}
	\label{fig:visulization_LINEo1o2}
}
\caption{Visualization of Co-author Network.}
\label{fig:visulization_3d}
\end{figure*}

\subsubsection{Ranking Evaluation}
The multi-relational network embedding approaches, such as TransE\cite{transe} and TransH\cite{transh}, usually utilize the metric, i.e. $||h+r-t||$ where h, r and t denotes the representation vectors for head, relation and tail respectively, to select out a bench of candidates for the ranking in link prediction.
Unfortunately, the single-relational network embedding, which is discussed in the paper, usually utilizes binary-classifier to determine the results of link prediction as shown in Sec. \ref{sec:classification} as there is no metric directly available as a ranking criterion. Thus, a pool of candidates cannot be provided for some special tasks, e.g., aligning the users across social networks\cite{ijcai/LiuCLL16}. 

Alternatively, we used the probability of the existence for an edge, which can be implicitly computed by the network embedding model, to evaluate the ranking for candidate selection. And we use the records, which had never appeared in the network embedding training process, as the test set. Fig.~\ref{fig:DatasetPartition} depicts the training sets and test sets we used in each task (embedding, classification, ranking).

A pair of vertices is scored by tracking the optimization direction of each model, which are detailed in Table \ref{tab:criteria_score}. Technically, we utilized the inner product ($u_i^Tu_j$ ) of two vertex vectors as the scoring criteria since $\sigma(u_i^Tu_j)$ constitutes the main part of the probability of the existence for an edge and the sigmoid function is strictly monotonically increasing. Then, we ranked all pairs $(v_i, v_j), j=1,2,...,|V|$ for a given $v_i$ based on the score. We used precision\cite{Precision}, recall\cite{Precision} and Mean Average Precision (MAP) \cite{MAP} to evaluate the prediction performance.

Table~\ref{tab:link_prediction_ranking} shows the ranking performance for all models. Our models (GANE and GANE-O1) outperform others again in terms of all evaluation metrics. Surprisingly, GANE-O1 provides quite impressive prediction @1 whereas the other models present rather unpleasant results. 

Even if both IRGAN and GANEs are based on GAN models, GANEs constantly have better performance than IRGAN. Moreover, GANE and GANE-O1 converge rapidly in comparison with IRGAN. Fig.~\ref{fig:Training_Curves} illustrates P@3 along with the number of iterations increasing. This may be accounted for the application of Wasserstein-1 distance.






\subsection{Clustering}

\subsubsection{Visualization - A Qualitative Analysis}

We first investigated the quality of embedding representations in an intuitive way by visualization. PCA\cite{PCA} was adopted to facilitate dimension reduction. 
In this paper, we selected three components obtained by PCA to visualize vertices of the network in 3-D space. The resulting visualizations with different embedding models are illustrated in Fig.\ref{fig:visulization_3d}. Only the visualizations in GANE and GANE-O1 present a relatively clear pattern for the labeled vertices where the authors devoted to the same research area are clustered together. GANE performs the most favorable layout in terms of clear clustering pattern. LINE variations perform not well as they require a rather dense network for the model training.




\subsubsection{Quantitative Analysis}

We applied k-means \cite{kMeans} to cluster all vertices in the low-dimensional vector space and set the number of clusters as 3. We utilized majority vote to label the three clusters as : "data mining", "machine learning", or "computer vision". Then, we quantitatively computed the accuracy of the vertices being clustered for each cluster, which is defined as the proportion of the ``accurately'' clustered vertices to the size of the cluster.

Fig.\ref{fig:acc_category_Label_Classification} illustrates the clustering accuracy achieved by different models on each cluster. Again, GANE produces the best accuracy which is consistent with the visualization. We argue that GANE can effectively preserve the proximities among vertices in the low-dimensional space.

\vspace{12pt}

In summary, our GANEs (GANE and GANE-O1) achieve the best performance for both link prediction and clustering tasks which demonstrates the strength of the generative adversarial framework. The first-order proximity intentionally adopted in GANE-O1 does not help to significantly boost the embedding performance as seen from the comparison between GANE and GANE-O1. We think that purposely to preserve the handcrafted structures may lead the embedding to overlook other underlying latent complex structures in the network, as it is impossible for us to explore all structures in conventional methods. However, GANE may provide a way to explore the underlying structures as complete as possible by incorporating link predictions as a component of the generative adversarial framework.

\section{Conclusion}

This paper proposes a generative adversarial framework for network embedding. To the best of our knowledge, it is the first attempt to learn network embedding in a generative adversarial manner. 

We present three variations of solutions, including GANE which applies cosine similarity, GANE-O1 which preserves the first-order proximity, and GANE-O2 which tries to preserves the second-order proximity of the network in the low-dimensional embedded vector space. Wasserstein-1 distance is adopted to train the generator
with improved stability. We also prove that GANE-O2 has the same objective function as GANE-O1 when negative sampling is applied to simplify the training process in GANE-O2.

Experiments on link prediction and clustering tasks demonstrate that our models constantly outperform state-of-the-art solutions for network embedding. Moreover, our models are capable of performing the feature representation learning and link prediction simultaneously under the adversarial minimax game principles. The results also prove the feasibility and strength of the generative adversarial models to explore the underlying complex structures of networks.

In the future, we plan to study the application of generative adversarial framework into multi-relational network embedding. We also plan to gain more insight into the mechanisms that GANs can employ to direct the exploration and discovery of underlying complex structures in networks.




\bibliographystyle{ACM-Reference-Format}
\bibliography{sample-bibliography}

\end{document}